\documentclass[runningheads]{llncs}

\usepackage{eccv}

\usepackage{eccvabbrv}

\usepackage{graphicx}
\usepackage{subcaption}
\usepackage{booktabs}
\usepackage{resizegather}
\usepackage{bbold}
\usepackage{subcaption}
\usepackage{times}
\usepackage{helvet}
\usepackage{algorithm}
\usepackage{algorithmic}
\usepackage{courier}
\usepackage{xcolor}
\usepackage{graphicx}
\usepackage{multirow}
\usepackage{amssymb}
\usepackage{array}
\usepackage{enumitem}
\usepackage{makecell}
\usepackage{wrapfig}
\graphicspath{{./images/}}

\usepackage[accsupp]{axessibility}  

\usepackage{hyperref}

\usepackage{orcidlink}

\begin{document}


\title{ZeroDSS: Semantic Segmentation from \\ Class Names Alone}

\titlerunning{ZeroDSS: Semantic Segmentation from Class Names Alone}

\author{
Tian Ming \and
Marzi Heidari \and 
Yuhong Guo
}

\institute{School of Computer Science, Carleton University, Ottawa, Canada
}

\maketitle

\begin{abstract}

Semantic segmentation depends heavily on real images with pixel-level annotations, making dense supervision a major obstacle to deployment in new domains. Recent diffusion-based methods reduce annotation costs by generating synthetic image--mask pairs from text prompts and cross-attention maps. However, they often rely on dataset-aligned captions or image-specific textual supervision, and degrade when only category names are available. We study task-specific data-free semantic segmentation, where training uses generic pretrained generative priors but has access to no target-domain images, masks, or captions---only a predefined category set.
We introduce Zero-Data Semantic Segmentation (ZeroDSS), 
a feedback-driven approach that treats synthetic supervision as an adaptively controllable training distribution rather than a direct surrogate for real data. ZeroDSS constructs a semantically structured prompt distribution through LLM-guided compositional generation and adaptive class rebalancing, then converts the prompts into image--mask pairs using diffusion attention. 
Its feedback-driven distribution refinement consists of a two-phase closed-loop training procedure with a single loop-back step. In Phase I, a segmenter is trained using confidence-weighted supervision and dual-scale consistency. In Phase II, model feedback is used to adaptively target underperforming categories and refine diffusion-derived masks using model predictions. 
After this refined initial training, a teacher–student framework is utilized
	in the final phase to iteratively improve the fused supervision 
	and segmenter through exponential moving average (EMA) updates.
Using synthetic training data only, ZeroDSS achieves 35.0\% mIoU on MS COCO with DeepLabV3, and 66.3\% and 63.5\% mIoU on PASCAL VOC with DeepLabV3 and Mask2Former, respectively. 
These results show that explicitly constructing and iteratively refining the synthetic supervision distribution can substantially reduce the synthetic-to-real gap in dense prediction.

\end{abstract}
\section{Introduction}
Semantic segmentation has advanced substantially through high-capacity architectures and large datasets with dense pixel annotations. Modern systems obtain strong results on curated real-image corpora such as PASCAL VOC~\cite{voc2012} and MS COCO~\cite{coco}, with representative architectures including DeepLab~\cite{deeplabv3}. This progress remains tightly coupled to access to real images and their dense pixel-level labels. In privacy-restricted applications, proprietary data silos, or rapidly emerging categories, collecting and annotating such data can be costly or infeasible, making dense supervision a central obstacle to deployment.

Generative models offer an alternative source of supervision. GAN-based approaches such as DatasetGAN and BigDatasetGAN synthesize pixel labels with limited additional annotation~\cite{datasetgan,bigdatasetgan}, but are constrained by object-centric generation and limited compositional diversity. Text-to-image diffusion models provide a more scalable route: their token-level cross-attention maps encode spatial--semantic correspondence and can be converted into synthetic image--mask pairs~\cite{stable_diffusion,dataset_diffusion,diffumask}. Existing diffusion pipelines, however, usually depend on dataset-aligned prompts derived from real captions or caption surrogates to approximate the target distribution. When only class names are available, naive prompt sampling produces biased category frequencies, implausible combinations, and unstable attention masks, increasing both semantic and visual mismatch with real data.

We therefore study class-name-only data-free semantic segmentation. Training may use generic pretrained language and generative models as semantic priors, but receives no task-specific real images, pixel annotations, or dataset-aligned captions. This setting shifts the key question from how to imitate individual real examples to how to construct an effective synthetic training distribution without observing the target domain.

To address this problem, we introduce Zero-Data Semantic Segmentation (ZeroDSS), which treats synthetic supervision as a distribution to construct, optimize under, and iteratively refine 
based on feedback rather than a direct proxy for real data. LLM-guided compositional validation and adaptive class rebalancing first shape a structured prompt distribution, which is instantiated as image--mask pairs through diffusion attention. Phase I learns an initial segmenter with confidence-weighted supervision and dual-scale consistency regularization. 
Phase II uses model feedback to identify underperforming categories, generate targeted samples, and refine diffusion-derived masks using model predictions. 
The updated training data are then fed back into Phase I for further segmentation model training.
Phase III performs class-balanced teacher-student self-training to improve the segmenter: a fixed teacher supplies predictions for fusion with the original attention masks, while iterative EMA updates progressively improve the teacher without introducing real-domain supervision.

The key contributions of this work are summarized as follows:
\begin{itemize}[topsep=0pt]
\item We formulate class-name-only data-free segmentation around explicit construction and 
	adaptive refinement of the synthetic training distribution.

\item We develop LLM-guided compositional validation, adaptive class-frequency balancing, and segmentation-oriented prompting for structured synthetic supervision.

\item We combine confidence-weighted consistency training, feedback-driven category and mask refinement, and EMA teacher--student self-training.

\item ZeroDSS reaches 35.0\% mIoU on COCO and 66.3/63.5\% mIoU on VOC with DeepLabV3/Mask2Former, improving the Phase I--II models by 3.8, 4.6, and 3.2 points; the Phase I--II loop-back also supports novel category expansion.
\end{itemize}

\section{Related Work}

\subsection{Synthetic Data for Semantic Segmentation}

Synthetic data has long been used to reduce the cost of dense annotation. GAN-based methods such as DatasetGAN and BigDatasetGAN project lightweight segmentation heads onto generative feature spaces to obtain pixel supervision~\cite{datasetgan,bigdatasetgan}; related work extracts semantic structure from generative representations without explicit labels~\cite{voynov2020unsupervised,melas2021unsupervised}. These approaches establish the value of generative priors, but their object-centric biases restrict scene diversity and large-scale distribution control. Simulator-rendered datasets such as GTA5 and SYNTHIA instead provide exact pixel labels directly from graphics engines~\cite{gta5,synthia} and are widely used for synthetic-to-real adaptation~\cite{adaptsegnet,cycda,dacs,fda}. Their scene layouts, co-occurrence patterns, and class frequencies are predefined, however, and changing them generally requires modifying the simulator. Recent work shows that downstream segmentation depends not only on annotation quality but also on synthetic--real distribution alignment~\cite{diffalign,synthetic_imagenet_diffusion}. ZeroDSS addresses this issue by constructing and adapting the distribution through prompts rather than relying on a fixed generator or simulator.

\subsection{Diffusion-based Pixel-level Annotation}

Diffusion models support dense prediction in two distinct ways. SegDiff, DDP, DiffusionInst, and MaskDiff formulate semantic, instance, or panoptic segmentation as supervised denoising~\cite{segdiff,ddp,diffusioninst,maskdiff}; diffusion is the predictor itself, and training requires ground-truth masks. A separate line repurposes pretrained text-to-image models as synthetic supervision engines. Stable Diffusion exposes token-level cross-attention maps~\cite{stable_diffusion}, which DiffuMask and Dataset Diffusion aggregate into pixel-level pseudo-labels aligned with generated images~\cite{diffumask,dataset_diffusion}. This removes the need for ground-truth segmentation masks, but the prompts commonly come from real captions or caption surrogates. ZeroDSS follows the attention-derived supervision paradigm while replacing dataset-aligned text with an explicitly constructed class-name-only prompt distribution.

\subsection{Semi-supervised and Self-training Segmentation}

Pseudo-labeling~\cite{pseudolabel}, consistency regularization, and teacher--student methods such as Mean Teacher, FixMatch, and CPS~\cite{meanteacher,fixmatch,cps} exploit unlabeled data by converting model predictions into training targets. Segmentation-specific approaches including ST++, ReCo, and ClassMix further improve structural reliability through confidence selection, representation consistency, or class-aware mixing~\cite{stpp,reco,classmix}. These methods generally assume access to unlabeled real images from the target domain. ZeroDSS instead self-trains exclusively on generated images. Moreover, its teacher predictions do not simply replace the available labels: they are fused class-wise with the original diffusion attention masks and confidence-filtered before supervising a continuously trained student. The method therefore adopts iterative teacher--student refinement while preserving the class-name-only data-free protocol.

\subsection{Prompt Construction and Language-guided Data Generation}

Text prompts are the main control interface for text-to-image generation, so their formulation directly shapes the semantics and composition of synthetic images. Dataset Diffusion uses ground-truth captions or BLIP-generated surrogates to retain dataset-specific scene statistics~\cite{dataset_diffusion,blip}. Rule-based templates such as ``a photo of a \{class\}''~\cite{stable_diffusion,glide} require less information but offer limited control over co-occurrence, frequency, and object visibility. Grounded and compositional generation methods provide stronger spatial control~\cite{grounded_diffusion,open_vocab_diffusion}, but do not explicitly build a task-level distribution from class names alone. ZeroDSS uses an LLM~\cite{gpt4} as a semantic prior to validate category combinations, regulate class balance, and realize concise segmentation-oriented scenes. 
Prompting is treated as distribution construction rather than only instance-level description.

\begin{figure*}[t]
  \centering
 \includegraphics[width=\textwidth]{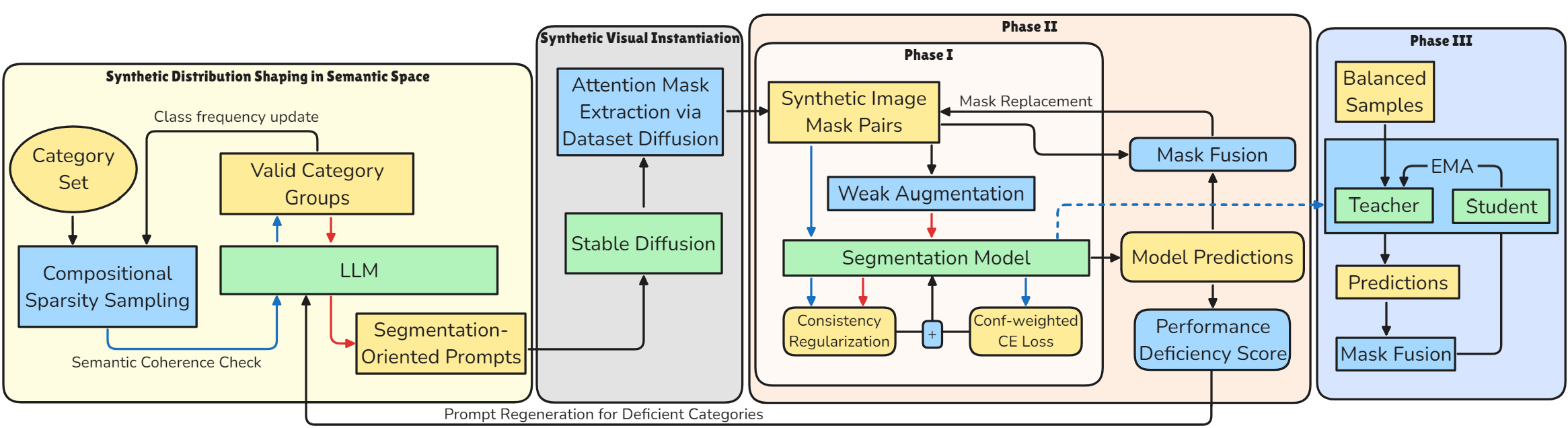}
 \vspace{0.8em}
\caption{
ZeroDSS constructs a structured synthetic distribution from class names through LLM-guided prompts and diffusion attention. Phase I trains with confidence-weighted supervision and dual-scale consistency. Phase II regenerates deficient categories, fuses diffusion masks with model predictions, and loops back to Phase I for continued training on the refined distribution. Phase III further enhance the model using a teacher-student framework, where the fixed teacher is updated by EMA between iterative rounds.
}
   \label{fig:approach}
\end{figure*}
\section{Problem Setting}
We study data-free semantic segmentation in which the only task-specific information available during training is a predefined foreground category set $\mathcal{C}=\{1,\ldots,C\}$; category $0$ is reserved for background. Unlike supervised or semi-supervised settings, training has access to no target-domain images, pixel-level annotations, or dataset-aligned captions. Generic pretrained language and generative models are allowed as reusable semantic priors, but they are not supplied with examples from the target segmentation dataset.
Starting from $\mathcal C$, LLM-generated prompts condition a text-to-image diffusion model to produce a synthetic dataset $\mathcal{D}^{\mathrm{syn}}=\{(\hat{\mathbf{x}}_i,\hat{\mathbf{m}}_i)\}_{i=1}^{N}$, where $\hat{\mathbf{x}}_i$ is a generated image and $\hat{\mathbf{m}}_i\in\{0,\ldots,C\}^{H\times W}$ is its attention-derived mask. The real evaluation set $\mathcal{D}^{\mathrm{real}}=\{(\mathbf{x}_j,\mathbf{m}_j)\}_{j=1}^{M}$ remains strictly unavailable during training. A segmenter $f_{\theta}:\mathbb{R}^{H\times W\times3}\rightarrow[0,1]^{H\times W\times(C+1)}$ assigns each pixel a probability vector over $C$ foreground classes and background. It is optimized only on $\mathcal{D}^{\mathrm{syn}}$ and evaluated only afterward on $\mathcal{D}^{\mathrm{real}}$.
The objective is therefore not merely to fit synthetic masks, but to generalize across a distribution shift that affects appearance, context, category co-occurrence, object scale, and mask reliability. Because the real distribution cannot be observed, ZeroDSS must improve both synthetic coverage and supervision quality using signals derived entirely from its generated data and trained segmenter.

\section{Method}

ZeroDSS treats synthetic supervision as an \emph{adaptivelly controllable training distribution} rather than a direct proxy for real data. Without real images, captions, or co-occurrence statistics, naive class-name prompting can produce compositional bias, severe class imbalance, and brittle attention masks. ZeroDSS therefore constructs a semantically structured prompt distribution with compositional and frequency control, then instantiates it into image--mask pairs through diffusion attention. 
Training proceeds through two phases of closed-loop refinement
with a single loop-back, followed by a third phase of model improvement.
Phase I learns a robust initial segmenter under noisy synthetic supervision, while Phase II uses model feedback to regenerate under-covered categories, refine their masks, and continue optimization on the updated distribution. 
Phase III iteratively improves fused pseudo-labels and segmenter
through EMA teacher--student self-training. Figure~\ref{fig:approach} summarizes the framework.

\subsection{Prompt Generation: Synthetic Distribution Shaping in Semantic Space}
Without target-domain category co-occurrence statistics, the synthetic distribution must be designed rather than inherited. Independently sampled categories can create implausible scenes and unstable attention supervision. ZeroDSS therefore uses an LLM as a semantic prior to control category composition, frequency, and prompt realization.

\paragraph{\bf Compositional Sparsity Control}
Natural images usually contain only a few salient categories. We model this sparsity by sampling the group size $k\sim\mathrm{Categorical}(r_1,r_2,r_3)$ for one-, two-, or three-category scenes, limiting implausible density and stabilizing attention masks while retaining multi-object diversity.
\paragraph{\bf Adaptive Category Frequency Rebalancing}
To counter category dominance, we maintain a frequency $n_c$ and sample category $c$ with 
the following weight
\begin{equation}
\label{eq:sample-weight}
w_c = \frac{1}{\left(n_c/\tau_c + \epsilon\right)^\alpha},
\end{equation}
where $\epsilon>0$ ensures numerical stability, $\alpha\in(0,1]$ controls rebalancing strength, and $\tau_c$ specifies a target frequency reflecting semantic centrality. Categories that naturally anchor scenes may receive larger $\tau_c$ without dominating the dataset. A group $g\subseteq\mathcal C$, $|g|=k$, is sampled without replacement in proportion to $\{w_c\}$, after which 
the frequency counts are updated: $n_c\leftarrow n_c+1$ for $c\in g$. 
Repeated updates gradually balance effective supervision while preserving diverse category combinations.

\paragraph{\bf Semantic Coherence Enforcement via Language Priors}
Frequency balance alone does not ensure scene plausibility. The LLM filters sampled groups according to whether their categories can naturally coexist in one spatio-temporal context and viewpoint. Retained groups form $\mathcal{G}=\{g_1,\ldots,g_N\}$, preventing balanced but logically inconsistent compositions from becoming supervision.

\subsubsection{Segmentation-Oriented Prompt Realization}
\begin{figure*}[t]
\centering
\setlength{\tabcolsep}{3pt}

\begin{tabular}{ccc}

\begin{tabular}{c}
\textbf{Category Group}\\person, skateboard\\[3pt]
\textbf{Prompt}\\A person jumping on a\\skateboard on a playground.\\[6pt]
\begin{tabular}{cc}
\includegraphics[width=0.12\linewidth]{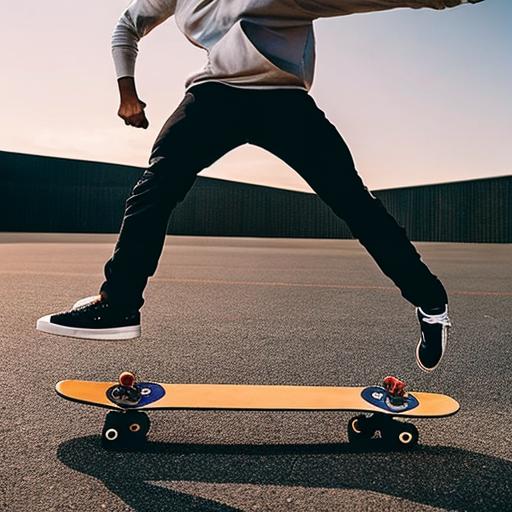} &
\includegraphics[width=0.12\linewidth]{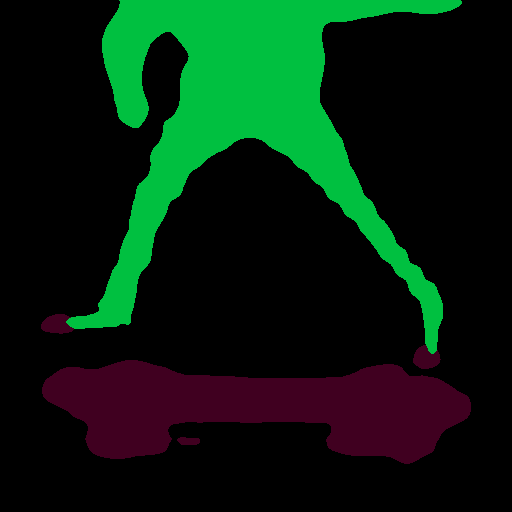}
\end{tabular}
\end{tabular}

&
\begin{tabular}{c}
\textbf{Category Group}\\clock, book\\[3pt]
\textbf{Prompt}\\A book in the bookshelf\\with one clock beside it.\\[6pt]
\begin{tabular}{cc}
\includegraphics[width=0.12\linewidth]{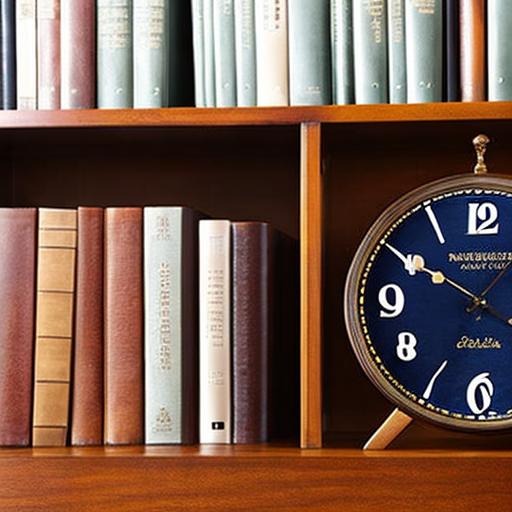} &
\includegraphics[width=0.12\linewidth]{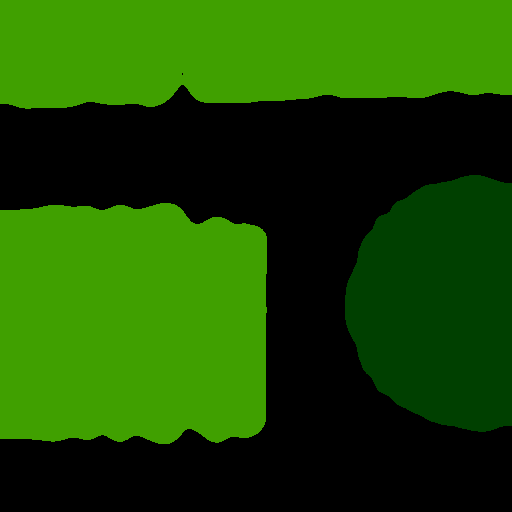}
\end{tabular}
\end{tabular}

&
\begin{tabular}{c}
\textbf{Category Group}\\pizza, knife\\[3pt]
\textbf{Prompt}\\A knife beside a pizza\\on a dining table.\\[6pt]
\begin{tabular}{cc}
\includegraphics[width=0.12\linewidth]{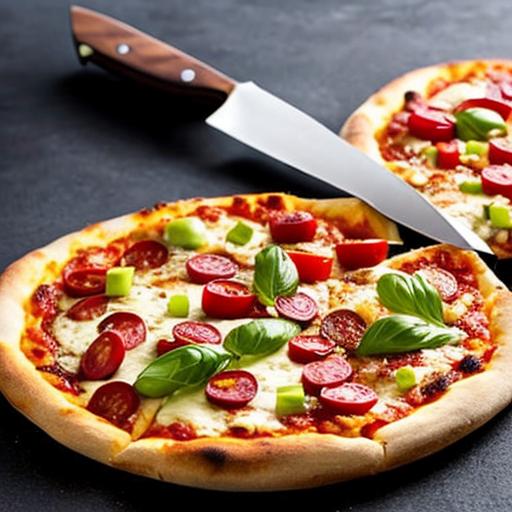} &
\includegraphics[width=0.12\linewidth]{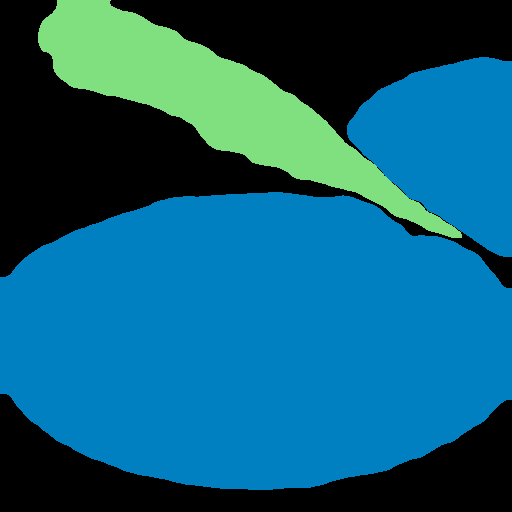}
\end{tabular}
\end{tabular}

\end{tabular}
\caption{Examples of segmentation-oriented prompts generated by the LLM and the corresponding attention-derived masks.}
\label{fig:prompt_examples}

\end{figure*}

For each $g_i\in\mathcal G$, the LLM $f_{\text{LLM}}$ generates a scene description $S_i$, which is concatenated with the category list to form a prompt:
\begin{equation}
	T_i=[S_i;g_i], \quad S_i = f_{\text{LLM}}(g_i, \mathcal{I}),
\label{eq:prompt_gen}
\end{equation}
where $\mathcal I$ is a structured instruction template
for prompt generation. Rather than requesting generic captions, it asks for short and literal scenes in which target objects are visible, spatially separable, and minimally occluded. Simple relations such as ``next to'' or ``on a table'' are preferred over crowded scenes, complex interactions, or descriptive background clutter. These constraints preserve semantic diversity while making diffusion cross-attention easier to convert into reliable masks. 
Examples of generated prompts are shown in Figure~\ref{fig:prompt_examples}.

To facilitate future synthetic mask generation with diffusion cross-attention, 
we also account for token-position bias in diffusion cross-attention
for segmentation-oriented prompt realization: 
Stable Diffusion often assigns stronger attention to objects mentioned earlier. 
Visually complex or segmentation-critical objects are therefore placed earlier when possible
in the prompts. The rule is applied conservatively; if reordering would produce unnatural phrasing or disrupt an interaction, the original relation is retained. This balances attention stability with scene coherence.

The resulting $\mathcal{T}_{\text{prompt}}=\{T_1,\ldots,T_N\}$ defines the semantic distribution used for training data synthesis.

\subsection{Data Generation: Synthetic Visual Instantiation}
\label{sec:instantiation}

The structured semantic distribution 
constructed through the prompts $\mathcal{T}_{\text{prompt}}$
is instantiated with the pretrained Stable Diffusion model $f_{\text{sd}}$. For each $T\in\mathcal T_{\text{prompt}}$, the model generates an image $\hat{\mathbf{x}}$, while cross-attention aggregation yields its aligned mask $\hat{\mathbf{m}}$~\cite{dataset_diffusion}. Denoting this process by $f_{\text{syn}}$ gives
\begin{align}
\mathcal{D}^{\mathrm{syn}} = \{(\hat{\mathbf{x}}_i, \hat{\mathbf{m}}_i)\}_{i=1}^N
	=f_{\text{syn}}(f_{\text{sd}},\mathcal{T}).
\end{align}

Our contribution lies in semantic distribution shaping rather than generative architecture modification. We therefore retain the mask extraction pipeline of Dataset Diffusion~\cite{dataset_diffusion}, isolating the effects of prompt construction and adaptive refinement from changes to the diffusion backbone.

\subsection{Feedback-Driven Synthetic Distribution Refinement}

ZeroDSS treats segmentation model optimization and synthetic data distribution refinement as a coupled process, aiming to promote generalizability. This closed-loop adaptive segmentation training framework involves two phases. Phase I learns a stable 
segmenter under noisy synthetic supervision. 
Phase II then targets underperforming categories and unreliable masks
to adaptively reshape the synthetic distribution.

\subsubsection{Phase I: Regularized Training under Synthetic Supervision}
Phase I trains $f_\theta$ exclusively on $\mathcal D^{\mathrm{syn}}$. Attention-derived masks may be incomplete, fragmented, or weakly aligned with object boundaries, making uniform empirical risk minimization sensitive to noisy pixels. We therefore use a confidence-weighted cross-entropy loss:
\begin{equation}
\label{eq:seg_loss}
\begin{aligned}
\mathcal{L}_{\mathrm{seg}}(\theta)
&=
\mathbb{E}_{(\hat{\mathbf{x}},\hat{\mathbf{m}}) \in \mathcal{D}^{\mathrm{syn}}}
\Bigg[
	\frac{1}{\sum_{u,v}\omega_{u,v}(\tilde{\mathbf{m}})}
\sum_{u,v}
	\omega_{u,v}(\tilde{\mathbf{m}})\,
\ell_{\mathrm{ce}}
\left(
\hat{\mathbf{m}}_{u,v},
f_{\theta}(\hat{\mathbf{x}})_{u,v}
\right)
\Bigg].
\end{aligned}
\end{equation}
%
Here, $\ell_{\mathrm{ce}}(\cdot,\cdot)$ denotes the standard cross-entropy loss function, 
and $\tilde{\mathbf m}$ is the normalized cross-attention response before discretization into $\hat{\mathbf m}$. 
The pixel-wise confidence weight $\omega_{u,v} \in [0,1]$
 is derived from the synthetic mask and 
 defined as the maximum response across foreground class categories:
\begin{equation} \label{eq:confidence} 
\omega_{u,v}(\tilde{\mathbf{m}})
=\max_{k\in\mathcal{C}}\;
\tilde{\mathbf{m}}_{[u,v,k]}.
\end{equation} 
The confidence weights modulate each pixel’s contribution to the overall loss, 
allowing uncertain or noisy regions in the synthetic mask $\hat{\mathbf{m}}$
to exert proportionally less influence on the optimization dynamics. 
This formulation treats synthetic supervision as soft guidance, 
 stabilizing early training dynamics.

Training only on generated images can overfit to diffusion-specific textures, lighting, and prompt-induced regularities. To encourage semantic rather than appearance-specific features, 
ZeroDSS enforces dual-scale consistency--pixel-level consistency and 
image-level consistency--to assist segmenter training. 

Let $\hat{\mathbf x}'$ be a weak, semantics-preserving augmentation of $\hat{\mathbf x}$. 
The pixel-level consistency
enforces local prediction stability under augmenting perturbation
via symmetric KL regularization: 
\begin{equation}
\label{eq:cons_pixel}
\begin{aligned}
\mathcal{L}_{\mathrm{cons}}^{\mathrm{pix}}(\theta)
&=\!
\mathbb{E}_{\hat{\mathbf{x}} \in \mathcal{D}^{\mathrm{syn}}}
\mathbb{E}_{u,v}\!
\frac{1}{2}
\Big[
\mathrm{KL}\!\left(f_{\theta}(\hat{\mathbf{x}})_{u,v} \| f_{\theta}(\hat{\mathbf{x}}')_{u,v}\right)
+
\mathrm{KL}\!\left(f_{\theta}(\hat{\mathbf{x}}')_{u,v} \| f_{\theta}(\hat{\mathbf{x}})_{u,v}\right)
\Big].
\end{aligned}
\end{equation}
This term penalizes spatially localized prediction drift at corresponding pixels. Because the divergence is symmetric, neither view is treated as a fixed target. Local agreement alone, however, does not prevent a global shift in predicted class proportions. 
The image-level consistency regularization term therefore aligns spatially averaged class distributions:
\begin{equation}
\label{eq:cons_global}
\begin{aligned}
\mathcal{L}_{\mathrm{cons}}^{\mathrm{img}}(\theta)
&=
\mathbb{E}_{\hat{\mathbf{x}} \in \mathcal{D}^{\mathrm{syn}}}\frac{1}{2}
\bigg[
	\mathrm{KL}\!\Big(\mathbb{E}_{u,v} (f_{\theta}(\hat{\mathbf{x}})_{u,v})\Big\|
	\mathbb{E}_{u,v} (f_{\theta}(\hat{\mathbf{x}}')_{u,v})\Big) \\
&\qquad\qquad\ \ \ +
	\mathrm{KL}\!\Big(\mathbb{E}_{u,v}(f_{\theta}(\hat{\mathbf{x}}')_{u,v})\Big\|
	\mathbb{E}_{u,v}(f_{\theta}(\hat{\mathbf{x}})_{u,v})\Big)
\bigg],
\end{aligned}
\end{equation}
where $\mathbb{E}_{u,v} (\cdot) = \frac{1}{HW}\sum_{u,v}(\cdot)$.

These two consistency terms are complementary: pixel consistency preserves local structure, whereas image consistency stabilizes the overall semantic composition of the prediction.
Together they form 
the dual-scale consistency loss
$\mathcal{L}_{\mathrm{cons}}(\theta)=
\mathcal{L}_{\mathrm{cons}}^{\mathrm{pix}}
+
\mathcal{L}_{\mathrm{cons}}^{\mathrm{img}}.
$
The overall training objective for Phase I is
\begin{equation}
\label{eq:total_loss}
\mathcal{L}(\theta)
=
\mathcal{L}_{\mathrm{seg}}
+
\lambda
\mathcal{L}_{\mathrm{cons}},
\end{equation}
where $\lambda>0$ controls the trade-off between cross-entropy and consistency regularization.
Together, confidence weighting limits the influence of noisy masks and consistency regularization discourages synthetic-specific solutions, yielding the Phase I checkpoint used for feedback-driven refinement.
%
%
\subsubsection{Phase II: Feedback-Driven Synthetic Distribution Refinement}

Although Phase I is trained on a balanced synthetic distribution, its construction remains heuristic: some categories can still lack sufficient scale, context, or appearance diversity. Phase II uses model feedback to identify these under-covered semantic regions
and refine the synthetic distribution. 
Let $\mathrm{IoU}_c$ be the class-wise IoU measure of the Phase I model for class $c$ on a held-out synthetic set $\mathcal D_{\mathrm{val}}^{\mathrm{syn}}$ generated in the same way as $\mathcal D^{\mathrm{syn}}$. 
We define a performance-deficiency score for class $c$ as
$r_c = (1 - \mathrm{IoU}_c) + \gamma \, \kappa_c$,
where $\kappa_c$ is a fixed LLM-derived category-level prior obtained by
grouping foreground categories into easy, average, and hard scale-difficulty levels and assigning the predefined scores, and
$\gamma$ denotes the trade-off hyperparameter. The first term captures
empirical segmentation error under the synthetic distribution, while the
second introduces a complementary prior on real-world scale-induced
difficulty.
The prior is a heuristic category-level cue, not a measured difficulty score from
real target-domain images.
The computed category deficiency scores provide feedback
to drive synthetic distribution refinement
through targeted deficient category expansion. 

\paragraph{Targeted Deficient Category Expansion.}
Let  $\mathcal{C}_{\mathrm{def}} \subseteq \mathcal{C}$  
denote the set of performance-deficient categories selected as the top-$K$ foreground categories with largest $r_c$, 
defining where the synthetic distribution should be expanded.
For the selected deficient categories, we employ LLM-guided prompt generation 
to enhance reliability and diversity. 
In particular, we generate new prompts using two complementary instruction templates to control scene complexity.
The first template encourages simple scenes with a single large object and minimal context, producing images where object boundaries are clearly visible. 
These samples provide reliable attention-derived masks and help the model learn accurate object contours and basic shape cues. 
The second template encourages more complex scenes involving contextual clutter, occlusion, multi-object interactions, and small-scale instances, better approximating realistic visual configurations in the target domain. 
Newly generated samples are then added into the synthetic dataset
to form an expanded training dataset  $\tilde{\mathcal D}^{\mathrm{syn}}$ for model retraining.
By expanding the performance-deficient categories, we 
redistribute the probability mass of the training data toward difficult semantic regions. 
This procedure can be viewed as distribution-level hard-example mining, 
operating in data space rather than feature space.
It increases support around model-identified weaknesses instead of perturbing intermediate features.

\paragraph{Mask Fusion and Continued Training under the Refined Distribution.}

The expanded training set $\tilde{\mathcal D}^{\mathrm{syn}}$ follows a refined distribution $\tilde p^{\mathrm{syn}}$. 
However, diffusion derived masks provide prompt-aligned localization but can omit valid regions or contain fragmented components, whereas the trained segmenter often predicts more coherent object structure. 
Therefore, for continued training on the refined distribution, we perform 
class-wise mask fusion to combine the complementary signals from the cross-attention supervision and 
segmentation model predictions, 
selecting the more reliable label and confidence for each class.
The fusion details are provided in the appendix. 
A single loop-back step
is then conducted by returning to Phase I for continued segmentation model training,
expecting to improve the segmenter 
with the refined training set and mask supervision.

\subsection{Teacher-Student Based Self-Training}
\label{sec:teacher_student}

In addition to the feedback-driven single loop-back refinement procedure, 
Phase III further enhances the segmenter using a teacher-student framework. 
It initializes teacher and student networks from the refined model obtained
from the single loop-back based two-phase refinement procedure, such as $\theta_T\leftarrow\theta^{\mathrm{II}}$ and $\theta_S\leftarrow\theta^{\mathrm{II}}$, 
but returns to the original diffusion derived attention-masks rather than the Phase II refined masks. 
The student network is updated using a sampled class-balanced subset in each round,
while the teacher remains fixed during the update of the student network, providing
stable pseudo-label targets for the student network.
Specifically, the teacher prediction and the original attention mask are fused 
in a class-wise manner to induce a pseudo-label $y_i$ for each pixel. 
Given the teacher and student prediction distributions $p_i^T$ and $p_i^S$
for pixel $i$, the label confidence $c_i$ is set as the largest probability
based on the teacher's prediction, $c_i=\max_k p_i^T(k)$. 
Only the confident pixels in $\Omega=\{i:c_i>\tau_{\mathrm{st}}\}$, determined 
by a threshold $\tau_{\mathrm{st}}=0.6$, are used for student network update:
\begin{equation}
\label{eq:self_training_loss}
\mathcal{L}_{\mathrm{st}}(\theta_S)
=-\frac{1}{|\Omega|}\sum_{i\in\Omega}c_i\log p_i^S(y_i).
\end{equation}
Diffusion attention and teacher predictions contribute complementary spatial evidence, while confidence filtering and weighting limit the influence of uncertain teacher regions. 
The teacher network is then updated in an EMA (exponential moving averaged) manner from the student model 
parameters between rounds: 
\begin{equation}
\label{eq:teacher_ema}
\theta_T \leftarrow \mu\theta_T+(1-\mu)\theta_S,
\end{equation}
where $\mu$ is the decay coefficient. Updating the teacher network only between rounds separates pseudo-label generation from student optimization; the new EMA teacher labels the next subset.
The final teacher network is returned as the model for evaluation. 

\section{Experiments}
\subsection{Experimental Setup}
\paragraph{Datasets.}
We evaluate ZeroDSS on PASCAL VOC 2012~\cite{voc2012} and MS COCO 2017~\cite{coco}, containing 20 and 80 foreground categories, respectively. Real images and annotations are strictly reserved for synthetic-to-real evaluation. Stable Diffusion v2.1-base~\cite{stable_diffusion}
generates approximately 40k VOC and 80k COCO image--mask pairs at $512\times512$ resolution with guidance scale 7.5. One-, two-, and three-category prompts are sampled with probabilities 
$(r_1,r_2,r_3)=(0.7,0.2,0.1)$. For frequency balancing, we set $\alpha=0.3$, $\epsilon=10^{-6}$, and $\tau_c=1$ except for \emph{person}, where $\tau_c=2$. We additionally generate 1k synthetic validation images per dataset for Phase II deficiency estimation.

\paragraph{Implementation Details.}
For synthetic data construction, prompts are generated using GPT-4.1 accessed via OpenRouter with moderate sampling temperature to encourage semantic diversity. 
Following prior synthetic segmentation baselines~\cite{dataset_diffusion}, we use DeepLabV3~\cite{deeplabv3} with a ResNet-50 backbone and no auxiliary head. AdamW uses learning rate $10^{-4}$, weight decay $0.01$, backbone multiplier $0.1$, a polynomial schedule with exponent $0.9$, batch size 8, and 40k Phase I iterations; the consistency coefficient is $\lambda=0.2$. Phase II fixes $\gamma=0.5$, selects the top $K=10$ deficient categories, generates 200 additional images per category. 
Phase III uses 10 rounds, approximately 25 images per foreground category per round, confidence threshold
is set as 0.6, and an EMA cosine schedule from 0.999 to 0.9999; the final EMA teacher is evaluated. 
Additional generation and training details are provided in the appendix.
All experiments report the results in terms of the mean Intersection over Union (mIoU).

\subsection{Comparison Results}

\begin{table*}[t]
\centering
\caption{
Comparison across supervision settings (mIoU). ZeroDSS uses approximately 82k COCO or 42k VOC synthetic training pairs after Phase II expansion and is evaluated on real validation sets. $^\dagger$ denotes reproduced results; open-vocabulary methods retain their original architectures.
}
\label{tab:main_comparison}
\resizebox{\textwidth}{!}{
\begin{tabular}{l l c c c}
\toprule
\textbf{Training Setting} &
\textbf{Method} &
\textbf{COCO(DLV3)} &
\textbf{VOC(DLV3)} &
\textbf{VOC(M2F)} \\
\midrule

	\textit{\makecell[l]{Fully Supervised Reference:\\ Real Images \& Annotations}}
& GT training
& 48.9 & 79.3 & 78.8 \\

\midrule

\textit{Open-Vocabulary Segmentation Models}
& CLIPSeg~\cite{clipSeg}
& 23.4 & \multicolumn{2}{c}{54.2} \\
& MaskCLIP~\cite{maskclip}
& 25.1 & \multicolumn{2}{c}{56.9} \\

\midrule

\textit{Synthetic Data with Real Dataset Captions}
& Dataset Diffusion$^\dagger$~\cite{dataset_diffusion}
& 28.6 & 58.0 & 57.8 \\
& DiffuMask~\cite{diffumask}
& -- & -- & 57.4 \\
& SDS~\cite{sds}
& 31.0 & 60.4 & 59.8 \\

\midrule

\textit{Synthetic Data with Class Names Only}
& CE baseline
& 26.2 & 57.4 & 57.6 \\
  & ZeroDSS (Phase I--II)
  & 31.2 & 61.7 & 60.3 \\
  & \textbf{ZeroDSS (full)}
  & \textbf{35.0} & \textbf{66.3} & \textbf{63.5} \\

\bottomrule
\end{tabular}}
\end{table*}

 We compare ZeroDSS against prior approaches from three training regimes: 
 real ground-truth supervision, prompt-aligned synthetic training, and open-vocabulary models. 
 Evaluation is conducted on MS COCO and PASCAL VOC using DeepLabV3 and Mask2Former backbones 
 pre-trained on ImageNet.
 All synthetic methods are trained solely on synthetic data 
 (80k samples for COCO and 40k for VOC) 
 and evaluated on the corresponding real validation sets. 
 Dataset Diffusion~\cite{dataset_diffusion}, DiffuMask~\cite{diffumask}, and SDS~\cite{sds} 
 are included as prompt-aligned synthetic baselines that use real dataset captions. 

 Table~\ref{tab:main_comparison} presents the main comparison results. 
 Under full real supervision, models trained on ground-truth annotations achieve 48.9\% mIoU on COCO and 79.3/78.8\% mIoU 
 on VOC using DeepLabV3 and Mask2Former respectively, establishing the upper bound for synthetic-to-real transfer. 
The foundational models,
CLIPSeg~\cite{clipSeg} and MaskCLIP~\cite{maskclip},
instead rely on strong open-vocabulary segmentation priors at inference and are included to 
contextualize the transfer enabled by pretrained vision-language representations. 
Among methods that train a conventional segmenter on synthetic pairs, 
SDS is the strongest baseline with dataset-caption supervision, reaching 31.0\% on COCO and 60.4/59.8\% on VOC,
 and outperforming Dataset Diffusion and DiffuMask. 
 DiffuMask is designed for single-object segmentation; therefore we report its VOC result with Mask2Former from the original paper.
 In contrast, ZeroDSS operates in the fully synthetic setting, 
 using only class names without access to real-image captions or dataset-specific semantic descriptions. 
 Despite this substantially weaker supervision signal, 
ZeroDSS, even with only the Phase I-II parts, 
reaches 31.2\% mIoU on COCO and 61.7/60.3\% on VOC with DeepLabV3/Mask2Former, 
exceeding the real-caption-aligned synthetic baselines despite operating with weaker task-specific information. 
With phase III, the full ZeroDSS further advances the performance to 
35.0\% and 66.3/63.5\% on COCO and VOC, respectively. 
This suggests the teacher predictions can provide structural cues complementary to diffusion attention, 
while class-wise mask fusion and confidence filtering reduce the effect of incomplete or uncertain masks during student optimization. 
The consistent gains across DeepLabV3 and Mask2Former backbones indicate that the refined supervision 
is not tied to a single segmentation architecture. 
Overall, ZeroDSS substantially improves synthetic-only segmentation and eliminates reliance on real-domain prompts, 
narrowing the gap to real supervision while maintaining a data-free training protocol.

\subsection{Ablation Studies}

\subsubsection{Ablation of the Learning Objective.}

Unless marked as the full model or Phase III study, all ablations, including novel category expansion, 
use the ZeroDSS variant with only Phase I-II. 
Table~\ref{tab:loss_ablation} reports the ablation results 
on COCO over the regularization terms of the learning objective. 
Without any regularization terms, 
the confidence-weighted segmentation loss $\mathcal L_{\mathrm{seg}}$ obtains a performance of 28.5\% in terms of mIoU. 
With image-level consistency regularization, the performance reaches 30.4\% (mIoU) by stabilizing the spatially averaged class distribution, while adding pixel-level consistency reaches 29.4\% by enforcing local invariance at corresponding locations. 
With both regularization terms, it achieves 31.2\% mIoU, confirming that global semantic composition and local boundary stability constrain different failure modes rather than duplicating one another.

\begin{table*}[t]
\centering
\begin{minipage}{0.48\linewidth}
\centering
\caption{Phase I--II loss ablation on COCO (mIoU).}
\begin{tabular}{ccc c}
\toprule
	$\mathcal{L}_{\mathrm{seg}}$ & $\mathcal{L}_{\mathrm{cons}}^{\mathrm{img}}$ & $\mathcal{L}_{\mathrm{cons}}^{\mathrm{pix}} $& mIoU (\%) \\
\midrule
$\checkmark$ &  &  & 28.5(+0) \\
$\checkmark$   & $\checkmark$ &  & 30.4(+1.9) \\
$\checkmark$   &  & $\checkmark$ & 29.4(+0.9) \\
$\checkmark$   & $\checkmark$ & $\checkmark$ & \textbf{31.2(+2.7)} \\
\bottomrule
\end{tabular}
\label{tab:loss_ablation}
\end{minipage}
\hfill
\begin{minipage}{0.48\linewidth}
\centering
\caption{Prompt construction ablation on COCO (mIoU). Val., Bal., and Aug. denote group validation, category balancing, and segmentation-oriented augmentation.}
\begin{tabular}{ccc c}
\toprule
	Val. & Bal. & Aug. & mIoU (\%) \\
\midrule
 &  &  & 26.6 \\
$\checkmark$ &  &  & 29.1 \\
$\checkmark$ & $\checkmark$ &  & 30.5 \\
$\checkmark$ & $\checkmark$ & $\checkmark$ & \textbf{31.2} \\
\bottomrule
\end{tabular}
\label{tab:grouping_ablation}
\end{minipage}
\end{table*}

\subsubsection{Ablation of LLM-guided Prompt Construction.}

 Table~\ref{tab:grouping_ablation} presents the ablation results of our LLM-guided prompt construction pipeline on COCO. 
LLM validation provides the largest gain (+2.5\% mIoU) by replacing random combinations with category groups that can plausibly coexist in one scene. This result supports using the language model as a semantic filter rather than merely as a caption generator. Without balancing, dominant classes such as \emph{person} exceed the next most frequent class by more than 10k instances; adaptive frequency control adds +1.4 points by redistributing effective supervision toward less frequent categories. Segmentation-oriented prompt augmentation with category list contributes a further +0.7 points by increasing contextual diversity while preserving visible, separable targets.

\subsubsection{Ablation of Phase III Mask Fusion--Pseudo-Label Strategy.}

\begin{wraptable}{r}{0.5\textwidth}
    \centering	
\centering
\small
\setlength{\tabcolsep}{5pt}
\caption{Effect of the Phase III pseudo-label source on COCO with DeepLabV3-R50. All variants evaluate the final EMA teacher.\\}
\label{tab:phase3_source_main}
\begin{tabular}{lc}
\toprule
	Pseudo-label source & mIoU(\%) \\
\midrule
Original attention mask only & 33.5 \\
Teacher prediction only & 34.4 \\
Class-wise fusion & \textbf{35.0} \\
\bottomrule
\end{tabular}
\end{wraptable}
We evaluate the effect of mask fusion in Phase III on COCO with DeepLabV3-R50, while keeping Phase I-II fixed.  
The class-wise fusion strategy adopted for Phase III produces pseudo-labels for the student network 
by fusing the diffusion derived attention mask with the teacher predictions. 
Table~\ref{tab:phase3_source_main} reports the ablation results 
by isolating these pseudo-label sources for Phase III. 
Original attention masks alone reach 33.5\% mIoU, whereas teacher predictions alone reach 34.4\%. 
Their class-wise fusion achieves 35.0\%, improving over the individual sources by 1.5 and 0.6 points. Diffusion attention retains prompt-aligned localization, while the teacher contributes more coherent structures learned by the segmenter; their complementary evidence explains the benefit of fusion.

\subsection{Effect of Mask Fusion and Prompt Refinement in Phase I-II}
We evaluate the impact of mask fusion and feedback-driven prompt regeneration in Table~\ref{tab:fuse_regen}
for the Phase I and II training. 
Applying mask fusion alone improves performance from 29.7\% to 30.8\% mIoU, indicating that refining the synthetic masks $\hat{\mathbf{m}}$ enhances supervision reliability by correcting structural artifacts before retraining. 
When regeneration over the performance-deficient category set $\mathcal{C}_{\mathrm{def}}$ is additionally introduced, performance further increases to 31.2 mIoU. 
This confirms that redistributing synthetic probability mass toward categories with high difficulty score $r_c$ improves overall segmentation quality.
Table~\ref{tab:regen_results} provides a class-wise analysis before and after refinement. 
While overall performance across $\mathcal{C}$ increases from 29.7\% to 31.2\% mIoU, the improvement is substantially larger for categories in $\mathcal{C}_{\mathrm{def}}$ (+3.25\% mIoU) than for those in $\mathcal{C}\setminus\mathcal{C}_{\mathrm{def}}$ (+1.27\% mIoU). 
This behavior is consistent with the design of the refinement stage. Regeneration explicitly increases the conditional sampling probability $\tilde{p}^{\mathrm{syn}}(c)$ for $c \in \mathcal{C}_{\mathrm{def}}$, thereby expanding synthetic support in underperforming semantic regions without degrading well-modeled categories.

\begin{table}[t]
\centering

\begin{minipage}{0.48\linewidth}
\centering
\caption{Effect of mask fusion and feedback-driven regeneration on COCO (mIoU).}
\begin{tabular}{cc c}
\toprule
Mask Fusion & Prompt Regeneration & mIoU \\
\midrule
 &  & 29.7 \\
$\checkmark$ &  & 30.8 \\
$\checkmark$ & $\checkmark$ & \textbf{31.2} \\
\bottomrule
\end{tabular}
\label{tab:fuse_regen}
\end{minipage}
\hfill
\begin{minipage}{0.48\linewidth}
\centering
\caption{Class-wise performance before and after refinement on COCO (mIoU).}
\begin{tabular}{lcc}
\toprule
Category Set & Before & After \\
\midrule
All Classes ($|\mathcal{C}|=81$) & 29.7 & 31.2 \\
Non-deficient ($\mathcal{C}\setminus\mathcal{C}_{\mathrm{def}}$) & 32.5 & 33.8 \\
Deficient ($\mathcal{C}_{\mathrm{def}}$) & 9.6 & 12.8 \\
\bottomrule
\end{tabular}
\label{tab:regen_results}
\end{minipage}

\end{table}
\vspace{-1em}

\subsection{Novel Category Expansion under Synthetic Supervision.}

\begin{table*}[t]
\centering
\setlength{\tabcolsep}{2pt}
\caption{
	Novel category expansion with ZeroDSS(Phase I-II) model (mIoU (\%)). Extra COCO categories are added to VOC without real training images.
}
\label{tab:voc_extension}
\resizebox{\linewidth}{!}{%
\begin{tabular}{c c c c  c c c c}
\toprule
\multicolumn{4}{c}{VOC + 5 Novel} & \multicolumn{4}{c}{VOC + 10 Novel} \\
\cmidrule(lr){1-4} \cmidrule(lr){5-8}
	Subset & Dataset Diffusion & CLIPSeg & \makecell{ZeroDSS\\(Phase I-II)} 
& Subset & Dataset Diffusion & CLIPSeg & \makecell{ZeroDSS\\(Phase I-II)} \\
\midrule
VOC (21) & 58.1 & 55.0 & 60.5
& VOC (21) & 57.5 & 55.8 & 59.9 \\

Novel(5) & 1.02 & 23.3 & 27.1
& Novel(10) & 0.38 & 21.9 & 25.9\\

All(26) & 47.1 & 49.0 & \textbf{54.7}
& All(31) & 39.1 & 44.9 & \textbf{48.9} \\
\bottomrule
\end{tabular}
}
\end{table*}

Table~\ref{tab:voc_extension} evaluates category expansion under synthetic supervision. 
Starting from the standard VOC category set, we introduce additional COCO categories during synthetic data generation without using any real images for the newly added classes. 
Performance is reported separately on the original VOC categories, the newly introduced categories, and the combined label space.

Dataset Diffusion maintains competitive performance on the original VOC categories but collapses on the newly introduced classes, suggesting limited category scalability beyond caption-aligned supervision. 
Because its generation process depends on dataset-specific image-caption statistics, it struggles to construct reliable supervision for categories absent from the original corpus.
CLIPSeg benefits from pretrained open-vocabulary segmentation priors and achieves substantially stronger performance on the novel categories. However, its performance on the original VOC categories remains below that of ZeroDSS.
In contrast, ZeroDSS directly incorporates new categories through structured prompt construction and synthetic segmentation training. 
It achieves 27.1\% and 25.9\% mIoU on the five- and ten-category extensions, respectively, while maintaining stable performance on the original VOC classes. These results suggest that structured synthetic supervision can support flexible category expansion under extremely weak task-specific supervision.


\section{Conclusion}
In this paper, we presented ZeroDSS for semantic segmentation in the class-name-only data-free setting. Rather than treating generated samples as a fixed surrogate for real data, 
the method constructs and progressively refines a synthetic supervision distribution
in a model feedback driven manner. 
LLM-guided prompt generation controls semantic composition and class frequency. 
The single loop-back based two-phase loop performs feedback-driven 
automatic synthetic distribution refinement and segmenter retraining.  
An additional Phase III further improves the model 
through class-balanced EMA teacher-student self-training.
Without target-domain images, masks, or captions, the full model reaches 35.0\% mIoU on COCO and 66.3/63.5\% on VOC with DeepLabV3/Mask2Former, improving the corresponding Phase I--II models by 3.8, 4.6, and 3.2 points. The Phase I--II model also supports novel category expansion while retaining performance on the original classes. These results indicate that synthetic-to-real transfer depends not only on generative image quality, but also on principled control of what is generated, how uncertain masks are used, and how supervision is updated during training.

\bibliographystyle{splncs04}
\bibliography{reference}
\newpage
\appendix

\section{Instructions for Prompt Generation}

The prompt-generation instruction is fixed throughout the synthetic data generation process and consists of several components designed to guide the large language model (LLM) toward producing segmentation-oriented captions. 
First, the instruction explicitly specifies the task objective and requirements, informing the LLM that the generated descriptions will be used for diffusion-based image synthesis for semantic segmentation. 
Second, it encourages realistic scene descriptions so that the prompts reflect natural visual situations rather than imaginative or fantasy-like compositions. 
Third, an attention-aware ordering strategy balances the placement of multiple objects in the sentence so that the diffusion cross-attention can focus more reliably on each target category. 
Fourth, lexical diversity is encouraged by using varied nouns for the same object types, which increases linguistic diversity without introducing additional prompt complexity. 
Finally, the instruction regulates the overall scene complexity, including object relations and background content, to avoid overly crowded compositions while maintaining natural interactions between objects.

During Phase II, additional prompts are generated with different levels of scene complexity to expand the training distribution of performance-deficient categories. 
Simple scenes emphasize clean spatial relations and minimal background clutter, which typically produce higher-quality cross-attention masks.
In contrast, complex scenes introduce multiple objects, richer spatial relations, and more detailed backgrounds. 
These images better approximate the visual diversity of real-world scenes and therefore improve generalization.
However, their cross-attention responses are often dispersed and therefore less confident. We retain these samples and their attention-derived masks during training: the confidence-weighted cross-entropy objective naturally assigns uncertain pixels less influence, without an explicit mask- or sample-level discard operation.

\begin{figure}[t]
  \centering
 \includegraphics[width=\linewidth]{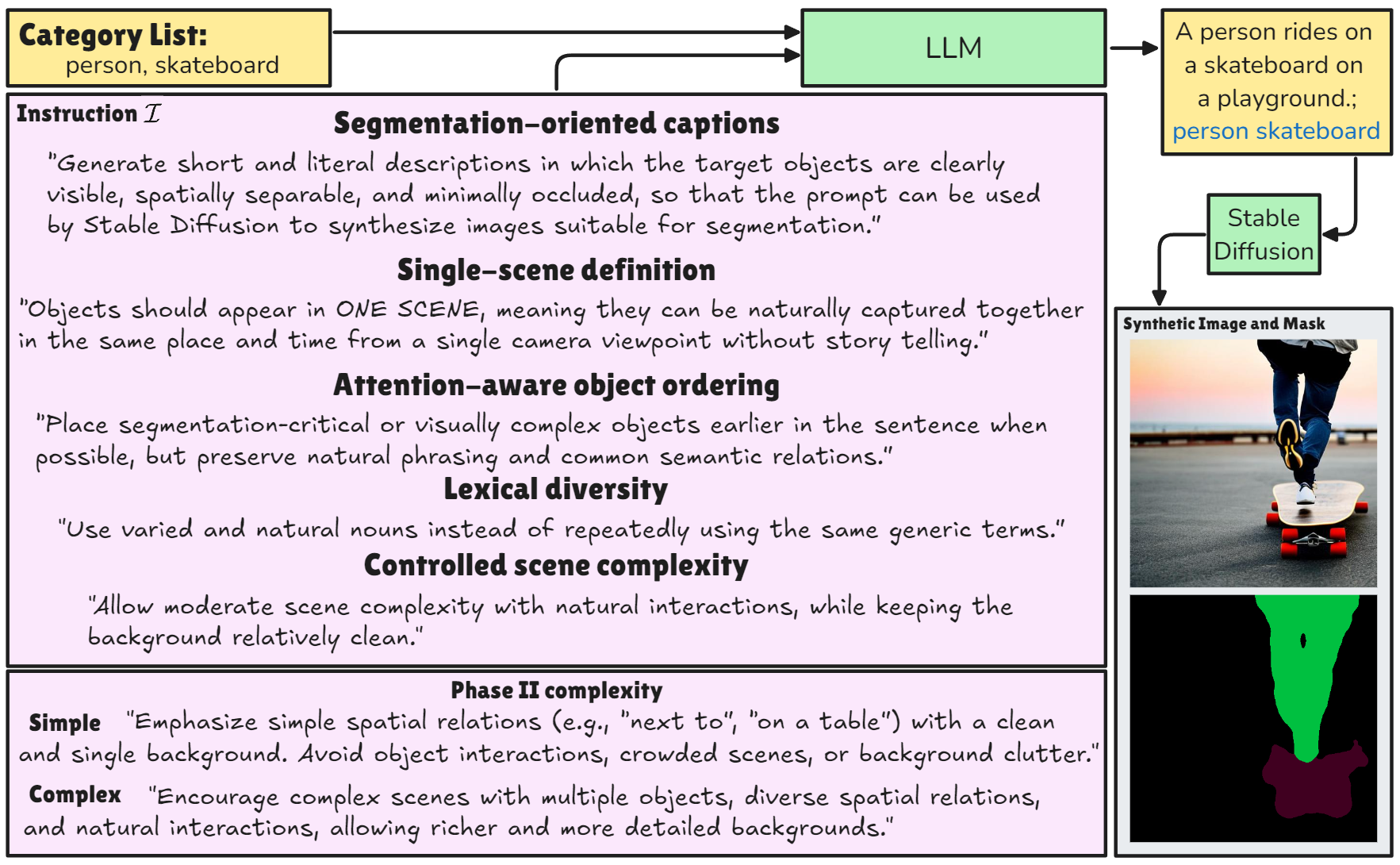} 
	 \caption{Structure of the instruction used for segmentation-oriented
prompt generation.}
   \label{fig:prompt_instruction}
\end{figure}

Figure~\ref{fig:prompt_instruction} presents the full instruction used for prompt generation, including the structured guidance described above and the two complexity templates used in Phase II.

\section{Qualitative Analysis of Prompt Construction}

\begin{figure*}[t]
    \centering
    \begin{subfigure}[t]{0.32\textwidth}
        \centering
        \includegraphics[width=0.48\textwidth]{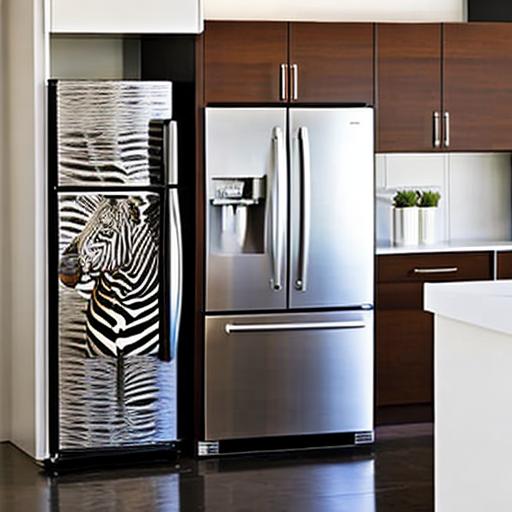}
        \hfill
        \includegraphics[width=0.48\textwidth]{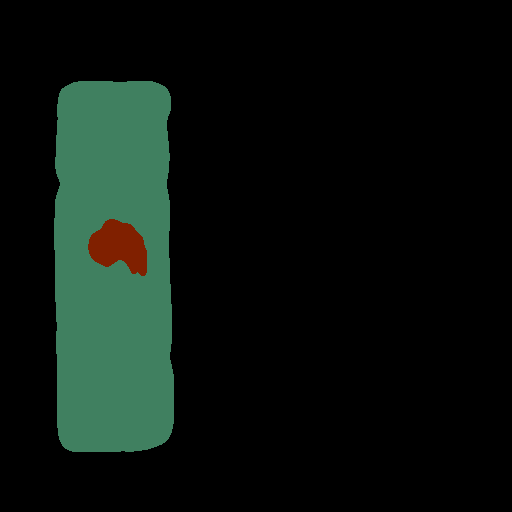}
        \vspace{0.8em}
        \caption{Random grouping: A \textbf{zebra} standing beside a \textbf{refrigerator} in a kitchen.}
    \end{subfigure}
    \hfill
    \begin{subfigure}[t]{0.32\textwidth}
        \centering
        \includegraphics[width=0.48\textwidth]{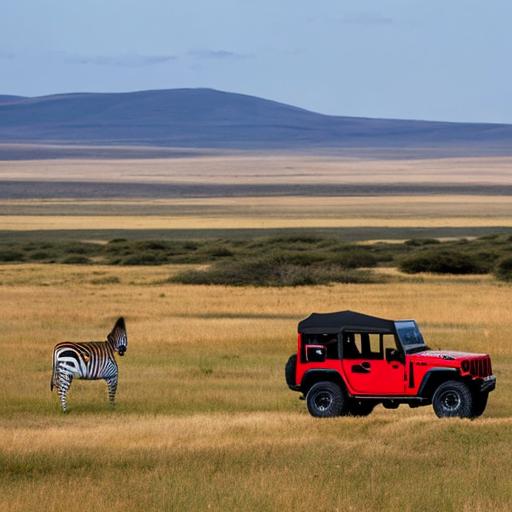}
        \hfill
        \includegraphics[width=0.48\textwidth]{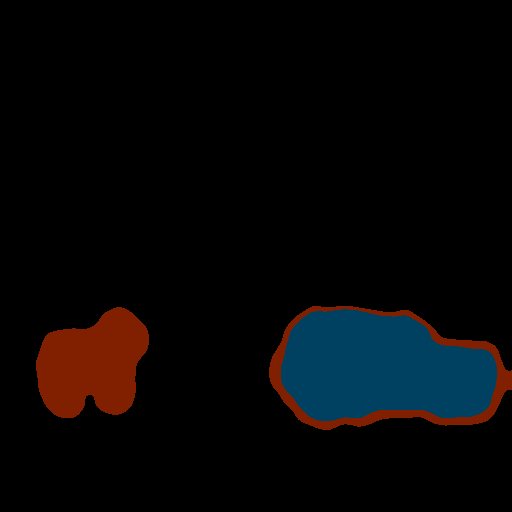}
        \vspace{0.8em}
        \caption{LLM grouping: A red \textbf{jeep} parked next to a \textbf{zebra} on a grassland.}
    \end{subfigure}
    \hfill
    \begin{subfigure}[t]{0.32\textwidth}
        \centering
        \includegraphics[width=0.48\textwidth]{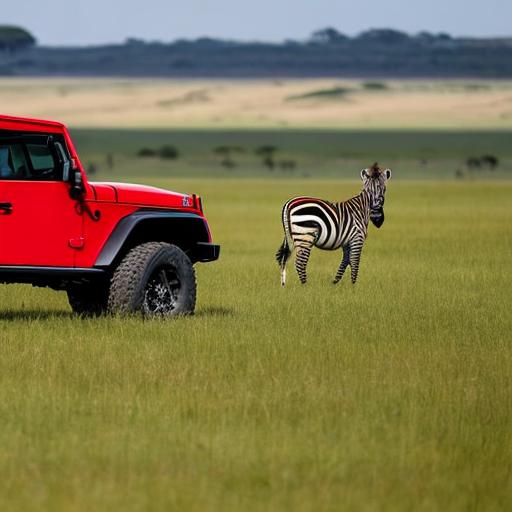}
        \hfill
        \includegraphics[width=0.48\textwidth]{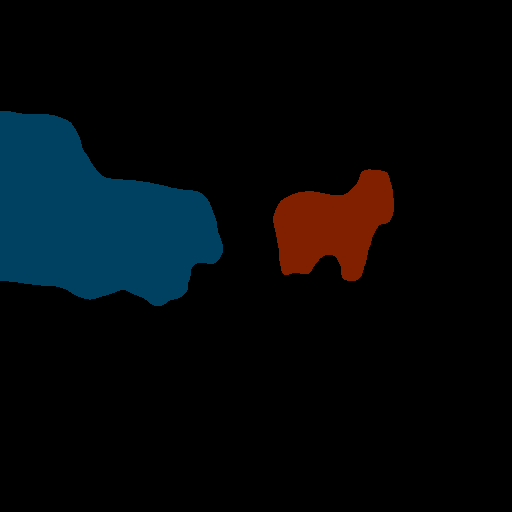}
        \vspace{0.8em}
        \caption{LLM grouping + augmentation: A \textbf{zebra} on the grassland next to a red \textbf{jeep}.}
    \end{subfigure}
    \caption{Effect of semantic grouping and token order on diffusion-based image generation and attention-derived masks.}
    \label{fig:grouping_compare_supp}
\end{figure*}

Figure~\ref{fig:grouping_compare_supp} compares random category grouping with LLM-validated grouping and segmentation-oriented prompt augmentation. Random grouping can produce implausible co-occurrences, whereas semantic validation yields more coherent object combinations and spatial layouts. 
Prompt augmentation with categories further improves token ordering and descriptive clarity, facilitating diffusion generation and producing cleaner attention-derived masks. These qualitative trends agree with the quantitative prompt-construction ablation reported in the main paper.

\begin{figure}[t]
\centering
\includegraphics[width=\linewidth]{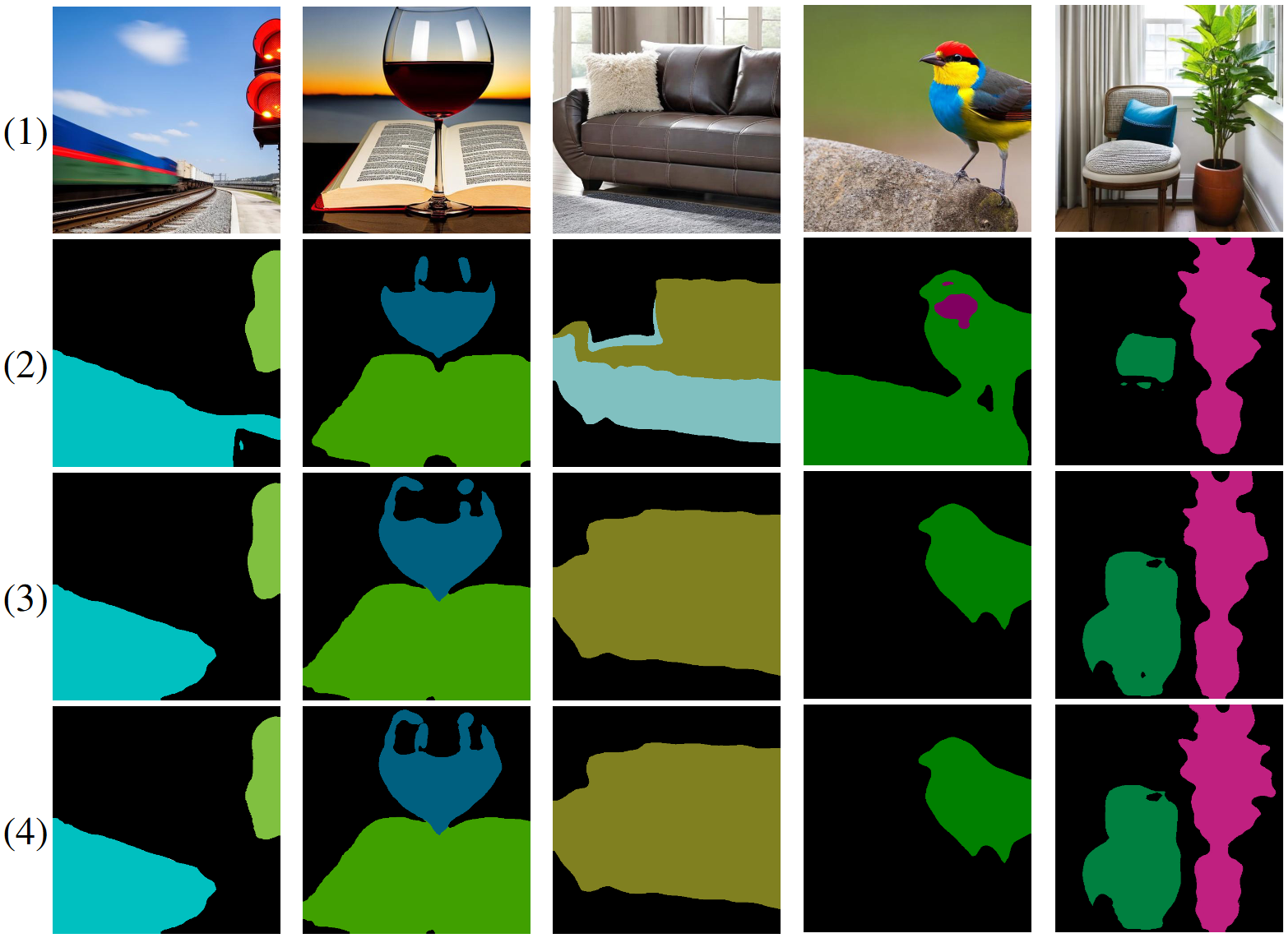} 
\caption{
Qualitative examples of mask refinement across four stages:
(1) synthetic image,
(2) diffusion cross-attention mask,
(3) segmentation model prediction,
and (4) fused mask after class-wise refinement.
Each column corresponds to one example.
}
\label{fig:fusion_example}
\end{figure}

\section{Class-wise Mask Fusion}
\label{sec:mask_fusion}
To improve the reliability of synthetic supervision, we refine diffusion-derived masks using segmentation model predictions in a class-wise manner. Diffusion attention often provides accurate category-conditioned localization but may omit valid regions or contain fragmented components, whereas model predictions usually produce more coherent object structures. 
We produce refined masks by fusing the diffusion-derived masks and the model predicted masks
in the following way. 
For category $c$, let $M_c^{\mathrm{att}}$ and $M_c^{\mathrm{pred}}$ denote the diffusion-derived mask and model-predicted mask, respectively. Negligible isolated components are removed from both masks before fusion.

Let $\mathfrak{C}(M_c)$ be the set of connected components in $M_c$, and let $\tau_{\mathrm{isl}}$ denote the island-size threshold. We define the structural-island indicator as
\begin{equation}
\label{eq:structural_island_supp}
\psi(M_c)=\mathbf{1}\!\left(\exists R\in\mathfrak{C}(M_c)\ \text{s.t.}\ |R|<\tau_{\mathrm{isl}}\right),
\end{equation}
where $\psi(M_c)=1$ indicates that the mask contains at least one structural island. 
Given this definition, the fused mask is set as
\begin{equation}
\label{eq:mask_fusion_supp}
M_c^{\mathrm{fus}}=
\begin{cases}
M_c^{\mathrm{att}},
	& \mbox{if}\;\psi(M_c^{\mathrm{att}})=0,\ \psi(M_c^{\mathrm{pred}})=1,\\
M_c^{\mathrm{pred}},
& \mbox{if}\; \psi(M_c^{\mathrm{att}})=1,\ \psi(M_c^{\mathrm{pred}})=0,\\
M_c^{\mathrm{att}}\cup M_c^{\mathrm{pred}},
& \mbox{if}\; \psi(M_c^{\mathrm{att}})=\psi(M_c^{\mathrm{pred}}).
\end{cases}
\end{equation}
Thus, when only one mask contains structural islands, the island-free mask is retained; when both masks have the same structural status, their union preserves complementary object regions. Categories appearing only in the model prediction are removed to avoid introducing hallucinated classes. Ambiguous or discarded pixels are assigned to the \textit{ignore} label, followed by light post-processing to improve spatial coherence. In Phase II, $M_c^{\mathrm{pred}}$ is produced by the Phase I model for newly generated samples. In Phase III, it is produced by the current teacher and fused with the corresponding original diffusion attention mask.

Figure~\ref{fig:fusion_example} provides qualitative examples of the mask refinement process. Common failure modes of diffusion-derived masks include fragmented activations, missing object parts, and isolated background regions. In many cases, segmentation model predictions provide more coherent object structures, while diffusion masks preserve reliable localization cues. The fusion strategy combines these complementary strengths to produce more complete masks.
The train and wine glass examples show recovery of missing regions and removal of isolated artifacts. The sofa and bird examples demonstrate the suppression of regions inconsistent with the generated image content. The chair example illustrates how small holes and fragmented regions can be filled after refinement.
Despite these improvements, some challenging structures remain difficult to recover. Extremely thin or low-contrast object parts, such as bird legs or wine-glass stems, may still be partially missing, and ambiguous attention activations can occasionally lead to incorrect region merging.

\begin{algorithm}[t]
\caption{\textbf{LLM-Guided Prompt and Synthetic Data Generation}}
\label{alg:data_generation}
\begin{algorithmic}[1]

\STATE \textbf{Input}: Category set $\mathcal{C}$; prompt instruction $\text{I}$; diffusion model $D$; target sample number $N$
\STATE \textbf{Output}: Synthetic dataset $\mathcal{D}^{\mathrm{syn}}$

\STATE Initialize valid group set $\mathcal{G}_{\mathrm{valid}}=\emptyset$

\WHILE{$|\mathcal{G}_{\mathrm{valid}}| < N$}
    \STATE Sample candidate group $g$ from $\mathcal{C}$
    \STATE Verify the semantic plausibility of $g$ using the LLM and obtain $v$
    \IF{$v = 1$}
        \STATE $\mathcal{G}_{\mathrm{valid}} \leftarrow \mathcal{G}_{\mathrm{valid}} \cup \{g\}$
    \ENDIF
\ENDWHILE

\STATE Initialize $\mathcal{D}^{\mathrm{syn}}=\emptyset$

\FOR{each group $g \in \mathcal{G}_{\mathrm{valid}}$}
    \STATE Generate prompt $T$ using LLM with instruction $\text{I}$
    \STATE Generate synthetic image $\hat{\mathbf{x}}$ using diffusion model
    \STATE Extract cross-attention maps from the diffusion process
    \STATE Convert attention maps into segmentation mask $\hat{\mathbf{m}}$
    \STATE $\mathcal{D}^{\mathrm{syn}} \leftarrow \mathcal{D}^{\mathrm{syn}} \cup \{(\hat{\mathbf{x}},\hat{\mathbf{m}})\}$
\ENDFOR

\RETURN $\mathcal{D}^{\mathrm{syn}}$

\end{algorithmic}
\end{algorithm}

\section{Algorithm for LLM-Guided Data Generation Process}

Algorithm~\ref{alg:data_generation} summarizes the synthetic data generation pipeline of our LLM-guided framework. The process consists of two stages: valid category-group construction and synthetic data generation.

First, candidate category groups are sampled from the predefined category set $\mathcal{C}$. Each sampled group is validated by the LLM to ensure semantic plausibility, i.e., whether the categories can naturally co-exist within a single scene. Only groups that pass this verification
step are added to the valid group set $\mathcal{G}_{\mathrm{valid}}$. This procedure continues until the desired number of valid groups is obtained.

Next, for each valid group, the LLM generates a segmentation-oriented prompt using the fixed instruction $\mathcal{I}$ as in Eq.(\ref{eq:prompt_gen}). The prompt is then used by the diffusion model to synthesize an image. Cross-attention maps from the diffusion process are extracted and converted into pixel-level segmentation masks, producing paired synthetic samples $\{(\hat{\mathbf{x}}, \hat{\mathbf{m}})\}$. The resulting image–mask pairs are accumulated to form the synthetic training dataset $\mathcal{D}^{\mathrm{syn}}$.

The synthetic validation set used for Phase II refinement is also generated using the same pipeline described in Algorithm~\ref{alg:data_generation}.


\begin{algorithm}[t]
\caption{\textbf{ZeroDSS Phase I--II Adaptive Segmentation Training}}
\label{alg:zerodss_training}
\begin{algorithmic}[1]

\STATE \textbf{Input}: Synthetic dataset $\mathcal{D}^{\mathrm{syn}}$; synthetic validation set $\mathcal{D}^{\mathrm{syn}}_{\mathrm{val}}$; initial segmentation model $f_{\theta^0}$; consistency weight $\lambda$
\STATE \textbf{Output}: Trained model $f_\theta$

\STATE Initialize $\theta \leftarrow \theta^0$

\STATE \textbf{\# Phase I: Initial Segmentation Training}

\FOR{each minibatch $\mathcal{B}=\{(\hat{\mathbf{x}}_i,\hat{\mathbf{m}}_i)\}_{i=1}^{N_b} \in \mathcal{D}^{\mathrm{syn}}$}
    \STATE Compute segmentation loss $\mathcal{L}_{\mathrm{seg}}$
    \STATE Compute pixel consistency $\mathcal{L}_{\mathrm{cons}}^{\mathrm{pix}}$
    \STATE Compute global consistency $\mathcal{L}_{\mathrm{cons}}^{\mathrm{img}}$
    \STATE Compute total loss
    $
    \mathcal{L}(\theta)
    =
    \mathcal{L}_{\mathrm{seg}}
    +
    \lambda\left(
    \mathcal{L}_{\mathrm{cons}}^{\mathrm{pix}}
    +
    \mathcal{L}_{\mathrm{cons}}^{\mathrm{img}}
    \right)
    $
    \STATE Update $\theta$ using gradient descent
\ENDFOR

\STATE \textbf{\# Phase II: Feedback-Driven Dataset Refinement}

\STATE Compute deficiency scores $r_c = (1-\mathrm{IoU}_c) + \gamma \kappa_c$ on $\mathcal{D}^{\mathrm{syn}}_{\mathrm{val}}$

\STATE Select deficient categories $\mathcal{C}_{\mathrm{def}}$ (top-$K$ by $r_c$)

\STATE Generate additional synthetic samples $\tilde{\mathcal{D}}^{\mathrm{syn}}$ conditioned on $\mathcal{C}_{\mathrm{def}}$ using Alg.~\ref{alg:data_generation}

\STATE Update dataset:
$
\mathcal{D}^{\mathrm{syn}}
\leftarrow
\mathcal{D}^{\mathrm{syn}}
\cup
\tilde{\mathcal{D}}^{\mathrm{syn}}
$

\STATE Predict segmentation masks for samples in $\tilde{\mathcal{D}}^{\mathrm{syn}}$ using the Phase~I model

\STATE Refine masks via class-wise mask fusion (Sec.~\ref{sec:mask_fusion})

\STATE Replace masks in $\tilde{\mathcal{D}}^{\mathrm{syn}}$ with the refined masks

\STATE \textbf{\# Final Training on Refined Dataset}

\STATE Continue training $\theta$ from the Phase I checkpoint on the refined dataset
$\mathcal{D}^{\mathrm{syn}}$
using the same optimization procedure as Phase I

\RETURN $\theta$

\end{algorithmic}
\end{algorithm}
%
\begin{figure}[t]
    \centering
    \includegraphics[width=0.7\linewidth]{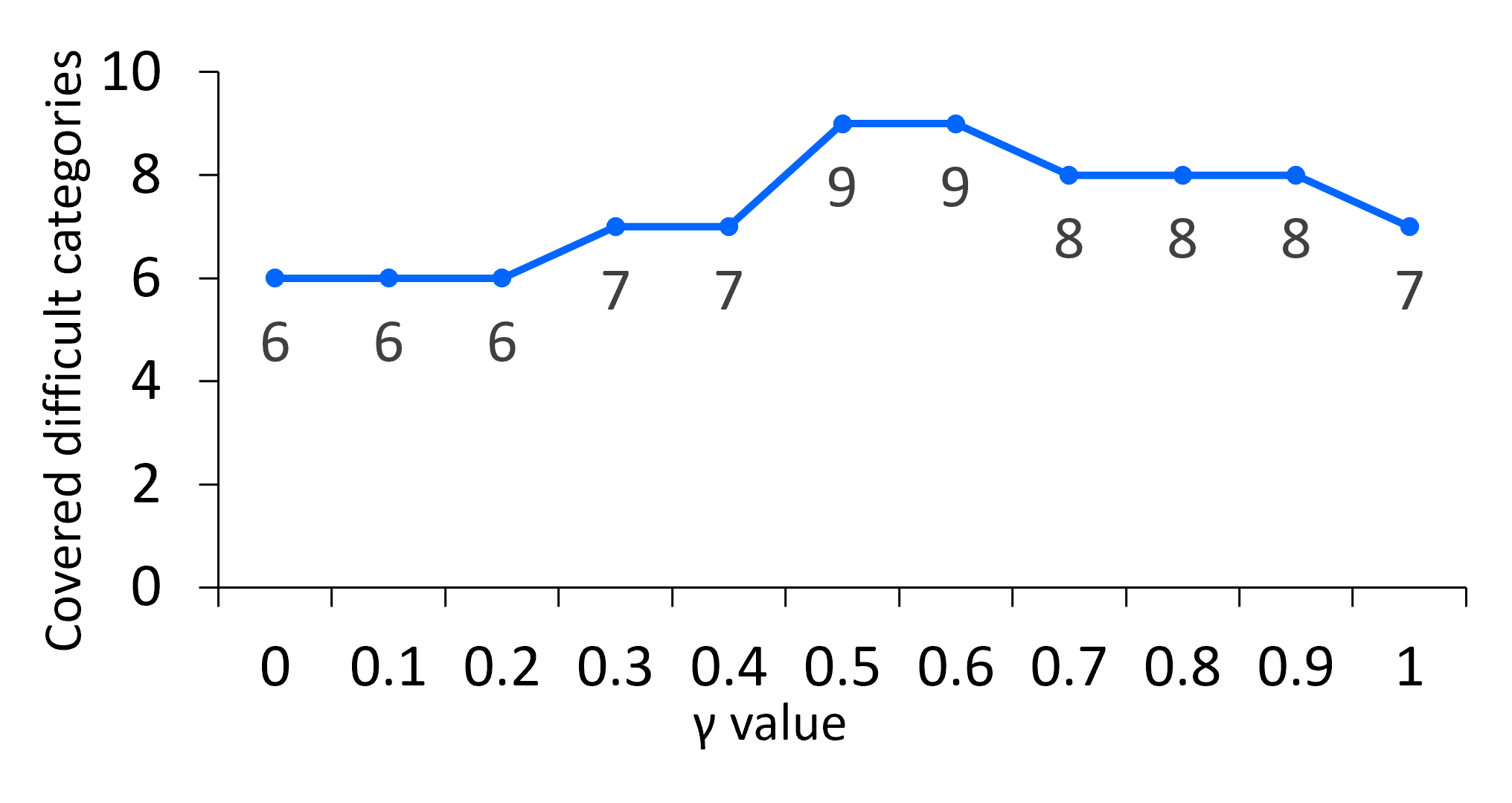}
    \caption{Effect of $\gamma$ on category selection quality.
The y-axis reports the number of selected top-$K$ deficient categories that correspond to below-average categories on the real COCO validation set.
Moderate values of $\gamma$ achieve the best alignment between synthetic-domain deficiency estimation and real-domain difficulty.}
    \label{fig:gamma_analysis}
\end{figure}
\begin{figure}[t]
    \centering
    \includegraphics[width=1\linewidth]{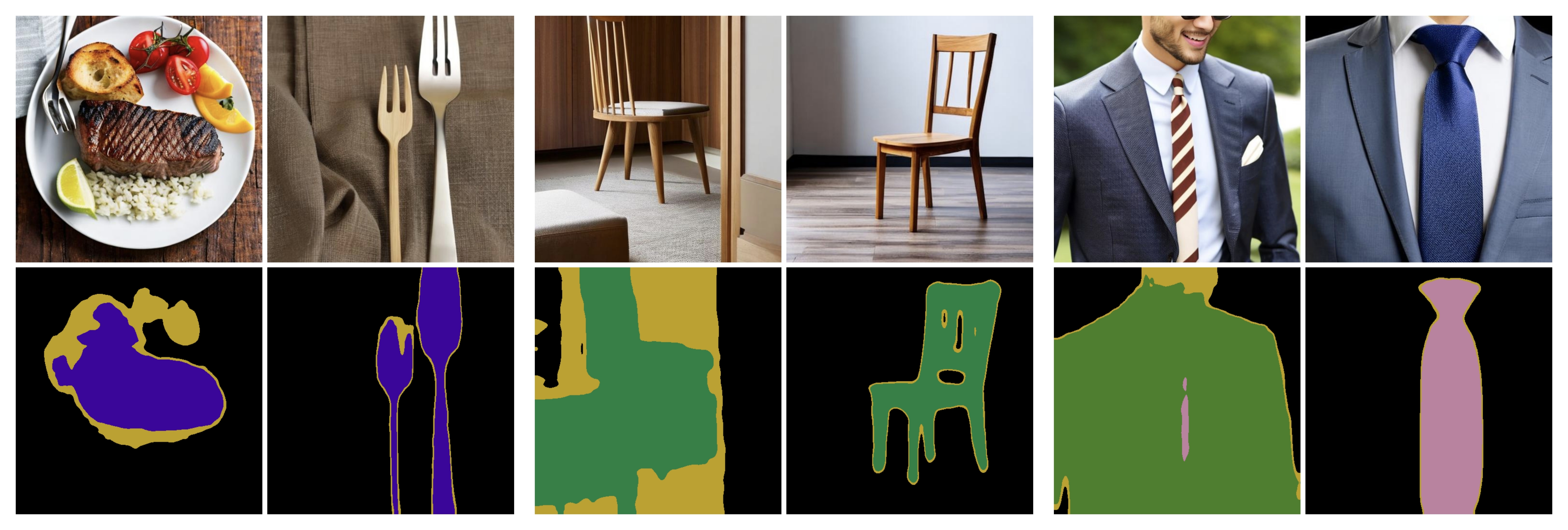}
    \caption{
	    Three examples of weak-category (fork, chair, necktie) regeneration in Phase II
    For each category, the left column shows the original synthetic sample and attention-derived mask, while the right column shows the regenerated sample and mask after adaptive refinement.
    }
    \label{fig:phase2_examples}
\end{figure}
%
\section{Algorithm for Adaptive Segmentation Training}

Algorithm~\ref{alg:zerodss_training} summarizes the Phase I--II adaptive training procedure of ZeroDSS. Starting from the synthetic dataset $\mathcal{D}^{\mathrm{syn}}$ generated by the diffusion pipeline, the segmentation model is first optimized using the confidence-weighted segmentation loss together with pixel-level and global consistency
regularization (Phase I).

The trained model is then evaluated on a synthetic validation set to estimate category-wise difficulty scores according to $r_c=(1-\mathrm{IoU}_c)+\gamma\kappa_c$. Since synthetic validation performance may not fully reflect real-domain difficulty, a lightweight category difficulty prior $\kappa_c$ is introduced as an auxiliary signal. Specifically, categories are assigned into three difficulty groups (easy, medium, and hard) using LLM semantic reasoning, corresponding to prior scores of 0, 0.5, and 1, respectively. The prior is used only for category ranking and does not directly influence segmentation model optimization.
The default $\gamma=0.5$ is fixed throughout all reported experiments. 
Categories with poor performance are identified as performance-deficient classes, and additional synthetic samples are generated to improve the training distribution (Phase II). The newly generated samples are further refined through class-wise mask fusion before being incorporated into the training set.

Finally, a single loop-back to Phase I is conducted, which continues training the segmenter
on the augmented synethetic dataset using the training procedure of Phase I.

To evaluate the influence of the difficulty prior, we perform a post-hoc sensitivity analysis by varying $\gamma$ while keeping all other components unchanged.
For each setting, we measure how many of the selected top-$K$ deficient categories correspond to below-average performing categories on the real COCO validation set.
As shown in Fig.~\ref{fig:gamma_analysis}, 
the category selection quality peaks at $\gamma=0.5$ and $\gamma=0.6$, indicating that a moderate contribution from the semantic prior helps compensate for bias in synthetic validation while avoiding over-reliance on prior knowledge. The fixed default therefore lies within the stable moderate range identified by this diagnostic analysis.

Figure~\ref{fig:phase2_examples} presents representative examples of weak-category regeneration during Phase II. Depending on the identified failure mode, the regeneration process simplifies or enriches the scene composition. The examples shown here correspond to the complexity-reduction branch, where the target object becomes more visually prominent and less affected by surrounding clutter. As a result, diffusion attention becomes more concentrated on the target category, producing cleaner and more complete object masks for segmenter training. 

\begin{algorithm}[t]
\caption{\textbf{ZeroDSS Phase III Teacher--Student Self-Training}}
\label{alg:teacher_student_training}
\begin{algorithmic}[1]
\STATE \textbf{Input}: Original synthetic image--attention-mask dataset $\mathcal{D}^{\mathrm{syn}}$; Phase I-II parameters $\theta^{\mathrm{II}}$; rounds $R=10$
\STATE \textbf{Settings}: Images per foreground category $q=25$; confidence threshold $\tau_{\mathrm{st}}=0.6$; EMA coefficients $\{\mu_r\}_{r=1}^{R}$
\STATE \textbf{Output}: Final EMA teacher $f_{\theta_T}$
\STATE Initialize $\theta_T\leftarrow\theta^{\mathrm{II}}$ and $\theta_S\leftarrow\theta^{\mathrm{II}}$
\FOR{$r=1$ to $R$}
    \STATE Sample a class-balanced subset $\mathcal{B}_r\subset\mathcal{D}^{\mathrm{syn}}$ with $q$ images per foreground category
    \STATE Freeze the teacher $f_{\theta_T}$ for the current round
    \STATE Predict teacher distributions $p^T=f_{\theta_T}(\hat{\mathbf{x}})$ for $\hat{\mathbf{x}}\in\mathcal{B}_r$
    \STATE Fuse teacher predictions with the corresponding original diffusion attention masks using class-wise mask fusion
    \STATE Set $c_i=\max_k p_i^T(k)$ and retain pixels $\Omega=\{i:c_i>\tau_{\mathrm{st}}\}$
    \STATE Optimize $\theta_S$ for exactly one epoch on the fused labels $y_i$ using confidence-weighted cross-entropy over $i\in\Omega$, without augmentation and with the Phase I optimization settings
    \STATE Update the teacher after the round: $\theta_T\leftarrow\mu_r\theta_T+(1-\mu_r)\theta_S$
\ENDFOR
\RETURN $f_{\theta_T}$
\end{algorithmic}
\end{algorithm}


\section{Teacher-Student Self-Training Algorithm}

Algorithm~\ref{alg:teacher_student_training} summarizes Phase III of ZeroDSS. Both networks are initialized from the final Phase I-II model. 
Each round uses a class-balanced subset drawn using the original attention-mask rather than the masks refined in Phase I-II. 
Approximately 25 images are sampled per foreground category, and every selected image appears once during the round. 
The teacher remains fixed while the student is trained for one epoch without augmentation, using the same batch size, optimizer, and learning rate as in Phase II. 
The confidence threshold and the finite 10-round budget limit the accumulation of uncertain pseudo-labels.
After each round, the teacher network is updated through EMA and its updated version generates predictions for the next round. No real-domain samples are introduced.
The EMA coefficient follows a cosine schedule from 0.999 to 0.9999. 
The final EMA teacher is used for evaluation. 

\section{Hyper-parameter Sensitivity Analysis}

\begin{figure}[t]
  \centering
 \includegraphics[scale=0.4]{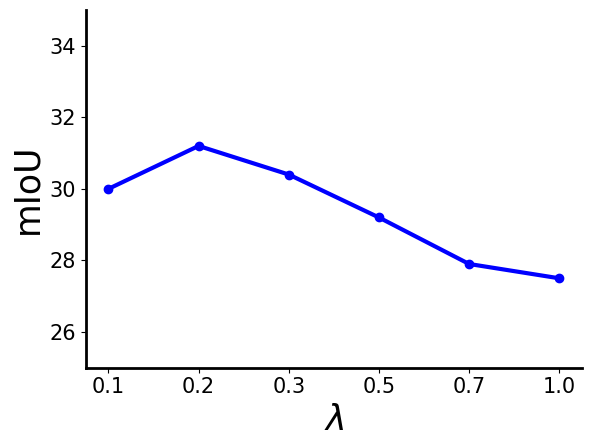} 
	 \caption{Sensitivity analysis of the consistency weight $\lambda$ on COCO (DeepLabV3-R50). 
}
   \label{fig:hyper-param}
\end{figure}

We analyze the effect of the consistency weighting coefficient $\lambda$
used in the training objective described in the main paper:
$\mathcal{L}(\theta)=\mathcal{L}_{\mathrm{seg}}+\lambda\mathcal{L}_{\mathrm{cons}}$. 
Figure~\ref{fig:hyper-param} presents the analysis results by varying the $\lambda$ value. 
Performance peaks at $\lambda=0.2$, indicating that a moderate contribution of the prediction-consistency term $\mathcal{L}_{\mathrm{cons}}$ is sufficient to stabilize training under synthetic supervision. 
When $\lambda$ is too small, the regularization effect is limited, reducing robustness to synthetic-specific artifacts. 
Conversely, larger values of $\lambda$ degrade performance, suggesting that excessive emphasis on $\mathcal{L}_{\mathrm{cons}}$ suppresses informative gradients from $\mathcal{L}_{\mathrm{seg}}$ and leads to underfitting. 
These results confirm that ZeroDSS benefits from balanced interaction between supervised risk minimization and distributional consistency regularization.

\end{document}